\documentclass[conference]{IEEEtran}
\makeatletter

\def\ps@IEEEtitlepagestyle{%
  \def\@oddfoot{}%
  \def\@evenfoot{}%
}

\usepackage{blindtext}
\usepackage{eso-pic}
\usepackage{cite}
\usepackage{amsmath,amssymb,amsfonts}
\usepackage{algorithmic}
\usepackage{graphicx}
\usepackage{textcomp}
\usepackage{accents}
\usepackage{xcolor}
\def\BibTeX{{\rm B\kern-.05em{\sc i\kern-.025em b}\kern-.08em
    T\kern-.1667em\lower.7ex\hbox{E}\kern-.125emX}}

\begin{document}

\title{The Classification of Optical Galaxy Morphology Using Unsupervised Learning Techniques}

\author{\IEEEauthorblockN{Ezra Fielding}
\IEEEauthorblockA{\textit{Department of Computer Science} \\
\textit{University of the Western Cape}\\
Cape Town, South Africa \\
0000-0002-7936-0222}
\and
\IEEEauthorblockN{Clement N. Nyirenda}
\IEEEauthorblockA{\textit{Department of Computer Science} \\
\textit{University of the Western Cape}\\
Cape Town, South Africa \\
0000-0002-4181-0478}
\and
\IEEEauthorblockN{Mattia Vaccari}
\IEEEauthorblockA{\textit{Department of Physics \& Astronomy} \\
\textit{University of the Western Cape}\\
Cape Town, South Africa \\
0000-0002-6748-0577}
}

\maketitle

\begin{abstract}
In recent years, large scale data intensive astronomical surveys have resulted in more detailed images being produced than scientists can manually classify. Even attempts to crowd-source this work will soon be outpaced by the large amount of data generated by modern surveys. This has brought into question the viability of human-based methods for classifying galaxy morphology. While supervised learning methods require datasets with existing labels, unsupervised learning techniques do not. Therefore, this paper implements unsupervised learning techniques to classify the Galaxy Zoo DECaLS dataset. A convolutional autoencoder feature extractor was trained and implemented. The resulting features were then clustered via k-means, fuzzy c-means and agglomerative clustering. These clusters were compared against the true volunteer classifications provided by the Galaxy Zoo DECaLS project. The best results, in general, were produced by the agglomerate clustering method. However, the increase in performance compared to k-means clustering was not significant considering the increase in clustering time. After undergoing the appropriate clustering algorithm optimizations, this approach could prove useful for classifying the better performing questions and could serve as the basis for a novel approach to generating more "human-like" galaxy morphology classifications from unsupervised techniques.
\end{abstract}


\section{Introduction}
The classification of galaxy morphology is an essential component in the investigation of galaxy formation and evolution \cite{walmsley2021galaxy}. Historically, this has been a process which scientists have performed by hand. However, with the advent of modern astronomical surveys, more detailed images are being produced than scientists can manually classify \cite{walmsley2021galaxy}. Projects such as Galaxy Zoo attempt to address this issue by collecting classifications from volunteers over the internet \cite{lintott2008galaxy}. So far, these projects have been largely successful, providing great benefits to the scientific community \cite{masters2019twelve}. However, the output of modern astronomical surveys will soon exceed human intelligence based efforts \cite{galvin2020cataloguing}. Discoveries and insights could be missed due to scientists being unable to process the massive volume of data produced by next generation surveys, such as the Square Kilometer Array \cite{dewdney2009square,fluke2020surveying}.

A variety of approaches have been used to address the problem of classifying galaxy morphology. The original Galaxy Zoo project, a human based approach, resulted in more than $4\times10^7$ individual classifications for nearly one million galaxies from the Sloan Digital Sky Survey (SDSS), collectively provided by approximately one hundred thousand volunteer participants \cite{lintott2008galaxy}. Subsequent projects, such as Galaxy Zoo DECaLS and Radio Galaxy Zoo, have also found success in making use of volunteers to inspect and visually classify galaxies \cite{banfield2015radio, walmsley2021galaxy}. The results of these projects have served as the basis for many galaxy morphology studies \cite{masters2019twelve}.

Despite the successes of these projects which rely solely on human classification and inspection, as the amount of survey data in need of classification increases, human based efforts will fail to keep up and require too much effort and time to remain viable \cite{galvin2020cataloguing}. It would have taken the recent Galaxy Zoo DECaLS project, with its 311,488 suitable galaxies, approximately 8 years to collect the standard 40 classifications per galaxy with the current number of volunteers \cite{walmsley2021galaxy}. This highlights the need for classification methods which do not rely heavily on human input.

For this reason, a significant amount of research has been done towards implementing supervised machine learning techniques to classify galaxy morphology. The use of artificial neural networks makes up a large portion of this research \cite{fluke2020surveying}. Recent literature has shown that, provided the sufficient availability of labeled datasets, convolutional neural networks (CNNs) are particularly accurate and useful in the classification of astronomical objects \cite{dieleman2015rotation, alhassan2018first, dominguez2018improving, burke2019deblending, lukic2018radio, ma2019machine, becker2021cnn, fielding2021comparison}. Bayesian CNNs have also been used to predict volunteer responses to decision trees to classify galaxies \cite{walmsley2021galaxy}.

However, these supervised machine learning methods still require labeled data in order to optimize and produce a usable model \cite{galvin2020cataloguing}. While there is no shortage of data in astronomy, relevant pre-labeled data is not always available \cite{fluke2020surveying}. This leaves supervised classification methods dependent on, and constrained by, the availability of human-generated labels. Recent work has been done on utilizing transfer learning to transfer existing models to new, unseen datasets, but it remains to be seen how successful these methods are and how well they scale \cite{tang2019transfer, galvin2019radio}.

Unsupervised machine learning is of particular interest concerning the problem of galaxy morphology classification, as it does not require existing labeled data. Instead, it identifies and classifies data by the structure of the data itself \cite{ghahramani2003unsupervised}. However, there has been significantly less work done on applying unsupervised machine learning algorithms to galaxy morphology classification when compared to the work done on supervised classification \cite{fluke2020surveying}. This suggests that there is a need for further research in this area. Notable works applying unsupervised learning to the domain of galaxy morphology classification include those done by Galvin \textit{et al} \cite{galvin2020cataloguing} and Ralph \textit{et al} \cite{ralph2019radio}. Ralph \textit{et al} propose the use of a convolutional autoencoder in combination with a self-organizing map to cluster radio astronomical survey data in an efficient and unsupervised manner \cite{ralph2019radio}. On the other hand, Galvin \textit{et al} propose the use of PINK, a rotationally-invariant self-organizing map algorithm, to associate radio and infrared sources without the need of existing training labels \cite{polsterer2019pink, galvin2020cataloguing}. Due to the unsupervised nature of these approaches, they operate without assumptions and labeled training data. This could result in unplanned, new and exciting discoveries in the world of astronomy \cite{ralph2019radio}. There has also been work done towards feature extraction on galaxy images, which also falls under the umbrella of unsupervised learning. Jiménez \textit{et al} propose feature extractor methods based on autoencoders and compare them to another state-of-the-art feature extractor called WND-CHARM \cite{jimenez2020galaxy}. The fact that these approaches, as with all unsupervised approaches, do not require existing labels, make them ideal to handle and process the large streams of data produced by next generation astronomical surveys.

The majority of existing work has focused on using unsupervised learning methods to make new and unscheduled discoveries without the need for existing labels. Therefore, this paper aims to determine the viability and merit of using unsupervised techniques to create clusters which conform to existing human generated galaxy morphology classifications. In other words, unsupervised methods will be tested to see how well they can reproduce human classifications given the same dataset and number of classes to produce. Features will be extracted from images used in the recent Galaxy Zoo DECaLS project using a convolutional autoencoder \cite{walmsley2021galaxy} . Three popular clustering methods will be considered, namely k-means clustering \cite{hartigan1979kmeans}, fuzzy c-means clustering \cite{ross2005fuzzy}, and agglomerative clustering \cite{Nielsen2016}. Clusters generated from extracted features will be mapped to existing real-world classes using the method proposed by Zha \textit{et al} \cite{zha2001spectral}. The three clustering algorithms will be compared based on how well the clusters they produce match the real-world classes. Accuracy metrics such as precision, recall and F1-score, as well as training time will be evaluated to determine clustering performance.

The remainder of this paper is organized as follows. A brief overview of the autoencoder feature extractor and clustering methods used in this work is provided in Section II. The materials and methods used in this work are presented in Section III, while Section IV presents the results and their discussion. A conclusion is drawn in Section V.

\section{Autoencoder and Clustering Methods Used in this Work}
This work makes use of a convolutional autoencoder as a feature extractor to reduce the dimensionality of the optical-spectrum galaxy image data. The clustering performance of three popular clustering methods were then tested using the extracted features. The methods considered were k-means clustering, fuzzy c-means clustering, and agglomerative clustering. The autoencoder and clustering methods are briefly described in the following subsections.

\subsection{Autoencoder Feature Extractor}
An autoencoder is an artificial neural network with a symmetric structure which is trained to reconstruct its input onto the output layer \cite{charte2018auto}. The output of the first half of the network represents an encoding of the input data. A convolutional autoencoder makes use of convolutional neural network mechanisms to complete this task. The convolutional autoencoder used in this work was based on the one proposed in the work of Jiménez \textit{et al} for extracting features from galaxy images \cite{jimenez2020galaxy}.

\begin{table}[htbp]
\caption{Autoencoder Model Summary}
\begin{center}
\begin{tabular}{|c|c|c|c|}
\hline
\textbf{Section}&\textbf{Layer (type)}&\textbf{Output} \textbf{Shape}&\textbf{Param \#}\\
\hline
\textbf{Encoder}&Conv2D&(None, 224, 224, 16)&448 \\
&MaxPooling2D&(None, 112, 112, 16)&0 \\
&Conv2D&(None, 112, 112, 8)&1160 \\
&MaxPooling2D&(None, 56, 56, 8)&0 \\
&Conv2D&(None, 56, 56, 8)&584 \\
&MaxPooling2D&(None, 28, 28, 8)&0 \\
&Flatten&(None, 6272)&0 \\
\hline
\textbf{Decoder}&Reshape&(None, 28, 28, 8)&0 \\
&UpSampling2D&(None, 56, 56, 8)&0 \\
&Conv2DTranspose&(None, 56, 56, 8)&584 \\
&UpSampling2D&(None, 112, 112, 8)&0 \\
&Conv2DTranspose&(None, 112, 112, 8)&584 \\
&UpSampling2D&(None, 224, 224, 8)&0 \\
&Conv2DTranspose&(None, 224, 224, 16)&1168 \\
&Conv2D&(None, 224, 224, 3)&435 \\
\hline
\end{tabular}
\label{tab:automodel}
\end{center}
\end{table}

Table~\ref{tab:automodel} presents the layer configuration for this project's implementation of the convolutional autoencoder. The autoencoder was split into encoder and decoder components, with the reverse structure of the encoder deployed for the decoder. Tensoflow 2.5 and Python 3.8 were used to implement this autoencoder \cite{tensorflow2015-whitepaper}. A rectified linear activation (ReLU) function was used along the network, except for the final output layer which applied the sigmoid activation function.

For convolutional autoencoders, the number of features extracted depends on the input image size \cite{jimenez2020galaxy}. In the case of the Galaxy Zoo DECaLS dataset, the color images (3 channels) which had been cropped to 224 by 224 pixels, resulted in an encoder which extracted 6272 features. Upon entry into the decoder portion of the network, the flattened feature vector was reshaped to the shape it held before flattening.

\subsection{K-means Clustering}
The standard scikit-learn implementation of the k-means algorithm was used in this paper \cite{scikit-learn}. The k-means clustering algorithm aims to partition $n$ objects into $k$ clusters in which each object belongs to the cluster with the nearest mean \cite{hartigan1979kmeans}. The algorithm minimizes the total intra-cluster variance, which is denoted as objective function $J$ and is defined by
\begin{equation}
    J = \sum_{j=1}^{k}\sum_{i=1}^{n}||x_{i}^{(j)}-c_{j}||^2
    \label{equ:kmeans}
\end{equation}
where $k$ is the number of clusters, $n$ is the number of data points, $x_{i}^{(j)}$ is the data point $i$, and $c_{j}$ is the centroid for cluster $j$.

The steps taken by this algorithm are as follows:
\begin{enumerate}
    \item Randomly choose cluster centers.
    \item Assign each object to the closest cluster center by Euclidean distance.
    \item Calculate the centroid by taking the mean of all objects in each cluster.
    \item Repeat until the same points are assigned to each cluster in consecutive rounds.
\end{enumerate}

\subsection{Fuzzy C-means Clustering}
The standard scikit-fuzzy implementation of the fuzzy c-means algorithm was used in this paper \cite{skfuzzy}. The objective function $J_m$ for a fuzzy c-partition matrix $\undertilde{\text{U}}$ for grouping a collection of $n$ data sets into  $c$ classes is defined by \cite{ross2005fuzzy}:
\begin{equation}
    J_{\text{m}}(\undertilde{\text{U}}, \text{v})=\sum_{k=1}^{n}\sum_{i=1}^{c}(\mu_{ik})^{m'}(\text{d}_{ik})^2,
    \label{equ:fuzzy1}
\end{equation}
where $\mu_{ik}$ is the membership of the $k$-th data point in the $i$th cluster and
\begin{equation}
    \text{d}_{ik}=\text{d}(\text{x}_{k}-\text{v}_{i})=\left[\sum_{j=1}^{m}(x_{kj}-v_{ij})^{2}\right]^{1/2}
    \label{equ:fuzzy2}
\end{equation}
denotes the distance measure between the $k$-th data point, $\text{x}_{k}$, and the $i$-th cluster center, $\text{v}_{i}$.

To gain the optimum fuzzy c-partitions, the objective function $J_m$ is minimized.
The steps taken by this algorithm are as follows:
\begin{enumerate}
    \item Set values for $c$, the number of classes, select a fuzzifier value, $m$, and initialise the partition matrix $\undertilde{\text{U}}^{(0)}$. Each step is labeled $r$, where $r$ = 0,1,2,...
    \item Calculate the $c$ centers $\{\text{v}_{i}^{(r)}\}$ for each step.
    \item Update the $r$th partition matrix $\undertilde{\text{U}}^{(r)}$.
    \item Stop if $||\undertilde{\text{U}}^{(r+1)}-\undertilde{\text{U}}^{(r)}||$ is less than or equal to the prescribed level of accuracy, otherwise increment $r$ and return to step 2.
\end{enumerate}

\subsection{Agglomerative Clustering}
The standard scikit-learn implementation of the agglomerative clustering algorithm was used in this paper \cite{scikit-learn}. Agglomerative clustering is a form of hierarchical clustering. The algorithm builds a binary merge tree starting from the leaves, which contain the data elements, to the root, which contains the full data-set \cite{Nielsen2016}. The Ward linkage function, which minimizes the sum of squared differences within all clusters, was used in this implementation. 

\section{Materials and Methods}
The following method was used for this paper. First a convolutional autoencoder was trained and then used to extract a number of features from the galaxy images. These extracted features were then used to cluster the galaxies using k-means clustering, fuzzy c-means clustering and agglomerative clustering for each GZD-5 decision tree question. The generated clusters were used to assign each galaxy to a class. Standard accuracy metrics were generated using these classes so that the results could be evaluated.

\subsection{Dataset Description}
The Galaxy Zoo DECaLS dataset includes 253,286 color images. These images are 424 by 424 pixels in size and feature 3 channels \cite{walmsley2021galaxy}. Volunteer responses are provided for each galaxy. These responses include the total votes and vote fractions for each question in the GZD-5 decision tree. Galaxies with less than 3 volunteer classifications were removed from the dataset leaving a total of 249,581 images. A random 80/20 train-test split was applied to the dataset. The result was a training dataset comprised of 199,664 galaxies and a test dataset comprised of 49,917 galaxies. The test dataset also served as the validation dataset used during autoencoder training, as well as the dataset used for clustering.

The GZD-5 decision tree features ten questions regarding the presence of specific characteristics found in the galaxy being evaluated. Each question has a different number of possible options to select \cite{walmsley2021galaxy}.

A set of discrete classifications for each GZD-5 decision tree question was produced by rounding the highest predicted volunteer vote fraction for a question to 1, while the remaining fractions were rounded to 0. This process was completed for each image. A galaxy was included for the evaluation of a decision tree question, only if 50 percent or more of the volunteers shown that galaxy answered that particular question.

In terms of pre-processing, galaxy images first had their pixel values normalized to fall between 0 and 1 and were cropped to 224 by 224 pixels. The color data for each image was preserved. A convolutional autoencoder was used to reduce the dimensionality of the image data and produce a latent feature vector for clustering purposes.

\subsection{Autoencoder Training and Testing}
The convolutional autoencoder was trained over 100 epochs with a batch size of 256 using the stochastic gradient descent (SGD) optimizer, in line with the method proposed by \cite{jimenez2020galaxy}. The mean squared error (MSE) loss function was used during training. Training for the autoencoder took approximately 10.6 hours to complete. Training was completed on a single NVIDIA V100 GPU.

Once the autoencoder was verified as having trained successfully, the 49,916 test images were fed into the encoder portion of the network for feature extraction. These features were then saved in a CSV file for later use.

\subsection{Clustering}
A set of 10 random seed values were used to ensure reproducible results for clustering algorithms which randomly generate initial states. All clustering was performed on a single compute node with access to 32 CPU cores.

\paragraph{K-means Clustering}K was set as the number of classes present for the question currently being evaluated. Centroids were randomly generated for each run and the max iterations parameter was set to 300. To account for random initialization, the k-means clustering algorithm was run 10 times for each question. The 10 sets of cluster assignments were used to calculate accuracy metrics.

\paragraph{Fuzzy C-means Clustering}C was set as the number of classes present for the question currently being evaluated. The fuzzifier value, m, was set to 2 and the convergence value was set to $1e-09$, in line with the input arguments proposed by \cite{cebeci2015comparison}. The initial fuzzy c-partition matrix was randomly set and the max iterations parameter was set to 300. To account for random initialization, the fuzzy c-means clustering algorithm was run 10 times for each question. The 10 sets of cluster assignments were used to calculate accuracy metrics.

\paragraph{Agglomerative Clustering}The number of clusters to be found is set as the number of classes present for the question currently being evaluated. The use of Ward linkage was specified and the remaining parameters were left at their default values as specified by scikit-learn \cite{scikit-learn}. The algorithm was only run once per question, as no random initialization was performed.

\subsection{Class Assignment}
The method proposed by Zha \textit{et al} was used to map each cluster to a corresponding 'real-world' class \cite{zha2001spectral}. First a confusion matrix was created using the existing cluster labels and true class labels generated from the Galaxy Zoo DECaLS discretized volunteer vote fractions. The sum of the diagonal of this matrix was then maximized by considering each possible matrix permutation and noting the configuration which results in the greatest sum across its diagonal. The permutation that produced the confusion matrix with the maximum sum of the diagonal was used to map each cluster to the 'real-world' class which it best fit.

\subsection{Generating Accuracy Metrics}
The mapped labels were used along with the true class labels to produce weighted average accuracy metrics for each question. These metrics included precision, recall and F1-score. See formulae \ref{equ:precision}, \ref{equ:recall} and \ref{equ:f1} for these definitions.
\begin{equation}
    \text{precision}=\frac{\text{True Positives}}{\text{True Positives}+\text{False Positives}}
    \label{equ:precision}
\end{equation}
\begin{equation}
    \text{recall}=\frac{\text{True Positives}}{\text{True Positives}+\text{False Negatives}}
    \label{equ:recall}
\end{equation}
\begin{equation}
    \text{F}_{1}=2\times\frac{\text{precision}\times\text{recall}}{\text{precision}+\text{recall}}
    \label{equ:f1}
\end{equation}

To account for label imbalance, the weighted average metrics were calculated for each question by taking the metrics for each question option, and finding their average weighted by support. Table~\ref{tab:support} shows the support for the metrics of each decision tree question. This represents the number of galaxies remaining for question evaluation in the test dataset after all galaxies which fell below the 50 percent volunteer threshold were removed.
\begin{table}[htbp]
\caption{Support by question}
\begin{center}
\begin{tabular}{|c|c|}
\hline
\textbf{Question Name}&{\textbf{Support}} \\
\hline
smooth or featured&49917 \\
disk edge on&15445 \\
has spiral arms&11380 \\
bar&11380 \\
bulge size&11380 \\
how rounded&32526 \\
edge on bulge&2475 \\
spiral winding&7499 \\
spiral arm count&7499 \\
merging&49247 \\
\hline
\end{tabular}
\label{tab:support}
\end{center}
\end{table}

\section{Results and Discussion}
The following section presents the weighted average accuracy metrics for each question produced by k-means, fuzzy c-means and agglomerative clustering. This is followed by a discussion of the results.
Table~\ref{tab:time} shows the average time in seconds taken to cluster each question by each clustering method.
\begin{table}[htbp]
\caption{Average Clustering Time (in Seconds)}
\begin{center}
\begin{tabular}{|c|c|c|c|}
\hline
\textbf{Question}&\multicolumn{3}{|c|}{\textbf{Clustering Method}} \\
\cline{2-4} 
\textbf{Name} & \textbf{\textit{K-means}}& \textbf{\textit{Fuzzy C-means}}& \textbf{\textit{Agglomerative}} \\
\hline
smooth or featured&41.4&231.9&6643.6\\
disk edge on&10.9&16.7&627.9\\
has spiral arms&8.4&12.5&339.8\\
bar&12.0&17.3&339.2\\
bulge size&16.2&24.9&339.2\\
how rounded&26.4&124.3&2792.4\\
edge on bulge&3.0&3.1&15.4\\
spiral winding&7.4&7.7&147.6\\
spiral arm count&12.0&12.5&147.5\\
merging&51.6&279.7&6384.5\\
\hline
\textbf{Total Time}&189.2&730.4&17777.3\\
\hline
\end{tabular}
\label{tab:time}
\end{center}
\end{table}

Tables~\ref{tab:precision}, \ref{tab:recall} and \ref{tab:f1} present a comparison of the precision, recall and F1-score, respectively, of the classes generated by each of the three clustering methods for each of the Galaxy Zoo DECaLS decision tree questions. The best performing results are presented in bold, while the worst performing results are underlined.
\begin{table}[htbp]
\caption{Precision by question}
\begin{center}
\begin{tabular}{|c|c|c|c|}
\hline
\textbf{Question}&\multicolumn{3}{|c|}{\textbf{Clustering Method}} \\
\cline{2-4} 
\textbf{Name} & \textbf{\textit{K-means}}& \textbf{\textit{Fuzzy C-means}}& \textbf{\textit{Agglomerative}} \\
\hline
smooth or featured&\textbf{0.592}&\underline{0.509}&0.528\\
disk edge on&\underline{0.654}&\textbf{0.711}&0.671\\
has spiral arms&0.728&\underline{0.714}&\textbf{0.736}\\
bar&0.445&\underline{0.385}&\textbf{0.460}\\
bulge size&\textbf{0.377}&0.303&\underline{0.266}\\
how rounded&0.435&\underline{0.422}&\textbf{0.457}\\
edge on bulge&0.608&\underline{0.457}&\textbf{0.687}\\
spiral winding&0.427&\underline{0.376}&\textbf{0.433}\\
spiral arm count&0.438&\underline{0.250}&\textbf{0.438}\\
merging&0.752&\underline{0.451}&\textbf{0.756}\\
\hline
\end{tabular}
\label{tab:precision}
\end{center}
\end{table}
\begin{table}[htbp]
\caption{Recall by question}
\begin{center}
\begin{tabular}{|c|c|c|c|}
\hline
\textbf{Question}&\multicolumn{3}{|c|}{\textbf{Clustering Method}} \\
\cline{2-4} 
\textbf{Name} & \textbf{\textit{K-means}}& \textbf{\textit{Fuzzy C-means}}& \textbf{\textit{Agglomerative}} \\
\hline
smooth or featured&\underline{0.395}&0.409&\textbf{0.497}\\
disk edge on&0.633&\underline{0.504}&\textbf{0.744}\\
has spiral arms&0.702&\underline{0.574}&\textbf{0.772}\\
bar&0.355&\textbf{0.418}&\underline{0.281}\\
bulge size&0.235&\textbf{0.384}&\underline{0.175}\\
how rounded&\textbf{0.387}&0.353&\underline{0.297}\\
edge on bulge&\textbf{0.427}&0.407&\underline{0.215}\\
spiral winding&\underline{0.384}&0.410&\textbf{0.470}\\
spiral arm count&0.197&\textbf{0.267}&\underline{0.189}\\
merging&0.262&\textbf{0.295}&\underline{0.232}\\
\hline
\end{tabular}
\label{tab:recall}
\end{center}
\end{table}
\begin{table}[htbp]
\caption{F1-Score by question}
\begin{center}
\begin{tabular}{|c|c|c|c|}
\hline
\textbf{Question}&\multicolumn{3}{|c|}{\textbf{Clustering Method}} \\
\cline{2-4} 
\textbf{Name} & \textbf{\textit{K-means}}& \textbf{\textit{Fuzzy C-means}}& \textbf{\textit{Agglomerative}} \\
\hline
smooth or featured&0.430&\underline{0.414}&\textbf{0.509}\\
disk edge on&0.643&\underline{0.545}&\textbf{0.696}\\
has spiral arms&0.714&\underline{0.618}&\textbf{0.749}\\
bar&0.361&\textbf{0.392}&\underline{0.250}\\
bulge size&0.251&\textbf{0.333}&\underline{0.202}\\
how rounded&\textbf{0.387}&0.364&\underline{0.298}\\
edge on bulge&\textbf{0.443}&0.400&\underline{0.236}\\
spiral winding&\underline{0.382}&0.384&\textbf{0.442}\\
spiral arm count&\textbf{0.241}&0.240&\underline{0.236}\\
merging&0.329&\textbf{0.332}&\underline{0.318}\\
\hline
\end{tabular}
\label{tab:f1}
\end{center}
\end{table}

As seen in Table~\ref{tab:precision}, agglomerative clustering generally resulted in the highest precision, while fuzzy c-means generally resulted in the lowest. It should be noted, however, that in most cases agglomerative clustering only narrowly outperformed k-means clustering. When considering the large discrepancy in run time between k-means clustering and agglomerative clustering, seen in Table~\ref{tab:time}, the precision performance gains of agglomerative clustering were not high enough to warrant the significantly longer training time.

Tables~\ref{tab:precision} and \ref{tab:recall} clearly depict a discrepancy between the precision and recall metric for each question for all three clustering methods. The precision was often significantly higher than the recall. This was best illustrated by considering the difference between precision and recall for the 'merging' question. Taking the definition of both precision and recall into account, one could conclude that the clustering methods considered resulted in a proportionally high amount of false negative classifications. This means that the produced clusters resulted in classes which excluded galaxies which should have been included, for each question considered. This could be a result amplified by the class assignment method used to map clusters to real world classes, described in the previous section.

When looking at these results broken down by question, it was clear that the accuracy performance for every clustering method considered was dependent on the nature of the question being considered. For example, higher precision was seen for questions with clear, distinct visual characteristics, such as 'disk edge on', 'has spiral arms', 'edge on bulge', and 'merging'. Questions with fewer possible classes seemed to also exhibit higher precision, provided the visual characteristics for each class were distinct enough. For questions with only two possible classes, in this case 'disk edge on' and 'has spiral arms', there was a smaller difference between the precision and recall, compared to the difference for other questions. This could indicate that, given distinct visual characteristics, the presence of fewer classes resulted in increased recall, however, further investigation will be required before a conclusion can be reached.

The F1-scores seen in Table~\ref{tab:f1}, provided good insight into the accuracy of the generated classes when both precision and recall were considered. For the 'merging' question, which had significantly higher precision than recall, the F1-score was low for all three clustering methods. While, on the other hand, the 'has spiral arms' question, which had a smaller difference between precision and recall, had higher scores for all three clustering methods.

Throughout testing and evaluation, it was clear that the random seed value used to initialize the k-means and fuzzy c-means functions heavily influenced the results and accuracy metrics produced. With this in mind, it appeared as though finding the optimal initial centroid positions and fuzzy c-partition matrix could result in better classification accuracy overall.

Based on the accuracy results seen in Tables~\ref{tab:precision}, \ref{tab:recall} and \ref{tab:f1}, and the behavior of the results observed during testing and evaluation, it seemed likely that, given the proper optimizations, this unsupervised approach could produce good results for the better performing questions. The insights produced by this implementation make it clear that unsupervised techniques are, in fact, useful in the field of classifying galaxy morphology. Even without comparing the clusters produced to existing labels, this implementation could result in new, unscheduled discoveries.

\section{Conclusion}
In conclusion, a convolutional autoencoder feature extractor was implemented and three clustering methods, namely k-means, fuzzy c-means, and agglomerative, were evaluated in terms of their accuracy and performance. Features were first extracted from the images in the Galaxy Zoo DECaLS dataset. The features were then clustered and the clusters formed were compared to existing human generated labels. Training time and accuracy metrics such as precision, recall and F1-score were used to evaluate performance. In terms of overall accuracy, agglomerative clustering appeared to have performed the best. However, the performance gain seen over k-means clustering was not significant considering the substantially longer clustering time required. The accuracy metrics produced proved useful for determining the best clustering method for each decision tree question. After undergoing the appropriate clustering algorithm optimizations, this approach could prove viable for classifying the better performing questions and could serve as the basis for a novel approach to generating more "human-like" galaxy morphology classifications from unsupervised techniques. This approach in its current form could still be used to make new and unscheduled discoveries from the clusters which are formed. Ultimately, the results of this investigation have shown that unsupervised learning techniques do, in fact, generate useful results and that they could be used to create clusters which conform to human generated galaxy morphology classifications.

\section*{Acknowledgment}
The authors acknowledge and thank the Galaxy Zoo volunteers who contributed to this dataset. Thanks is also given to Mike Walmsley for making the Zoobot Python library publicly available and for grating access to the Galaxy Zoo DECaLS dataset. The authors acknowledge the Telkom Center of Excelence at the University of the Western Cape for providing financial support, as well as the ilifu cloud computing facility for providing computational resources.

\bibliographystyle{IEEEtran}
\bibliography{references}

\end{document}